\documentclass{article}

\usepackage{arxiv}

\usepackage[utf8]{inputenc} 
\usepackage[T1]{fontenc}    
\usepackage{hyperref}       
\usepackage{url}            
\usepackage{booktabs}       
\usepackage{amsfonts}       
\usepackage{nicefrac}       
\usepackage{microtype}      
\usepackage{amsthm}
\usepackage{amsmath}
\usepackage{amssymb,dsfont,bbold,enumitem,mathtools}
\usepackage[dvipsnames]{xcolor} 
\usepackage{tikz}

\usepackage{caption}
\usepackage{wrapfig}

\usepackage{enumitem}

\usepackage{algorithmic}
\usepackage[ruled,linesnumbered]{algorithm2e}

\usepackage{hyperref}
\definecolor{mylinkcolor}{RGB}{0,0,140}
\hypersetup{colorlinks,allcolors=mylinkcolor,citecolor=mylinkcolor}

\newcommand{\RR}{\mathbb{R}} 
\def\one{\mathbb 1}
\def\tilde{\widetilde}
\def\bar{\overline}

\def\Diag{\mathrm{Diag}}
\def\ep{\varepsilon}
\def\algname{HyperND}
\newtheorem{theorem}{Theorem}[section]

\let\oldt\Theta
\def\Theta{\mathit \oldt}

\def\Phi{\mathcal L}

\usepackage{comment}

\usepackage[sort,numbers]{natbib}

\begin{document}

\title{A nonlinear diffusion method for semi-supervised learning on hypergraphs}

\author{%
     \textbf{Konstantin Prokopchik} \\
   School of Computer Science \\
   Gran Sasso Science Institute \\
   67100, L'Aquila Italy\\
   \texttt{konstantin.prokopchik@gssi.it}
   \And
   \textbf{Austin R. Benson} \\
   Department of Computer Science\\
   Cornell University\\
   Ithaca, NY 14853 \\
   \texttt{arb@cs.cornell.edu} 
   \And
   \textbf{Francesco Tudisco} \\
  School of Mathematics\\
  Gran Sasso Science Institute\\
  67100, L'Aquila Italy \\
  \texttt{francesco.tudisco@gssi.it} 
}

\maketitle

\begin{abstract}
Hypergraphs are a common model for multiway relationships in data, and hypergraph semi-supervised learning is the problem of assigning labels to all nodes in a hypergraph, given labels on just a few nodes. 
Diffusions and label spreading are classical techniques for semi-supervised learning in the graph setting, and there are some standard ways to extend them to hypergraphs.
However, these methods are linear models, and do not offer an obvious way of incorporating node features for making predictions.
Here, we develop a nonlinear diffusion process on hypergraphs that spreads both features and labels following the hypergraph structure. 
Even though the process is nonlinear, we show global convergence to a unique limiting point for a broad class of nonlinearities and we show that such limit is the global minimum of a new regularized semi-supervised learning loss function which aims at reducing a generalized form of variance of the nodes across the hyperedges.
The limiting point serves as a node embedding from which we make predictions with a linear model.
Our approach is competitive with state-of-the-art graph and hypergraph neural networks, and also takes less time to train.
\end{abstract}

\section{Introduction}
%
\label{sec:introduction}
In graph-based semi-supervised learning (SSL), one has labels on a small number of nodes,
and the goal is to predict labels for the remaining nodes. 
Diffusions, label spreading, and label propagation are classical techniques for this problem,
where known labels are diffused, spread, or propagated over the edges in a graph~\cite{zhou2004learning,zhu2003semi}.
These methods were originally developed for graphs where the set of nodes corresponds to
a point cloud, and edges are similarity measures such as $\epsilon$-nearest neighbors; however, they can also be used with relational data such as social networks or co-purchasing~\cite{chin2019decoupled,gleich2015using,juan2020ultra,kyng2015algorithms}.
In the latter case, diffusions are particularly well-suited  because they directly capture the idea of homophily~\cite{mcpherson2001birds} or assortativty~\cite{newman2002assortative},
where labels are smooth over the graph.

While graphs are widely-used models for relational data, many complex systems
and datasets are better described by higher-order relations that go beyond
pairwise interactions~\cite{battiston2020beyond,benson2018simplicial,torres2020why}.
For instance, co-authorship often involves more than two authors, people
in social network gather in small groups and not just pairs, and emails can have
several recipients. A hypergraph is a standard representation for such data,
where a hyperedge can connect any number of nodes.
Directly modeling  higher-order interactions has led to improvements
in several machine learning settings~\cite{zhou2007hypergraph,benson2016higher,li2017inhomogeneous,li2018submodular,yadati2019hypergcn,srinivasan2021learning}.

Along this line, there are a number of diffusions or Laplacian-like regularization techniques
for SSL on
hypergraphs~\cite{zhou2007hypergraph,hein2013total,zhang2017re,li2020quadratic,liu2020strongly,veldt2020minimizing,tudisco2021nonlinear},
which are also built on principles of similarity or assortativity.
These methods are based on the optimization of a convex Laplacian-based regularized loss which enforces local and global consistency across the hypergredges. This optimization formulation makes them easy to interpret. 
However, unlike the graph case, non-quadratic hypergraph consistency losses are typically required to model  real-world data interactions at  higher-order  \cite{chan2020generalizing,neuhauser2021consensus,tudisco2021node}. Thus, the corresponding diffusion algorithms, based on e.g.\ gradient flow integration, are not linear. 
Moreover, these methods are topically designed for cases where only labels are available, and do not take advantage of rich features or metadata associated with hypergraphs that are potentially useful for making accurate predictions.
%
For instance, co-authorship or email data could have rich textual information.

Graph and hypergraph neural network (GNN) is a popular approach that uses both features and network structure
for SSL~\cite{yadati2019hypergcn,feng2019hypergraph,dong2020hnhn}.
Hidden layers of GNNs aggregate the features of neighboring nodes via neural networks and learn the model parameters by fitting to the labeled nodes.
While combining features according to the hypergraph structure is a key idea, GNNs do not take direct advantage of the fact that connected nodes likely share similar labels. Moreover, they can be expensive to train.
In contrast, diffusion methods work precisely because of homophily and are typically fast.
In the simple case of graphs, combining these two ideas has led to several recent advances~\cite{klicpera2018predict,huang2020combining,jia2021unifying}.

Here, we combine the ideas of GNNs and diffusions for SSL on hypergraphs
with a method that simultaneously diffuses both labels and features according to the hypergraph structure.
In addition to incorporating features, our new diffusion can incorporate a broad class of nonlinearities to increase the modeling capability,
which is critical to the architectures of both graph and hypergraph neural networks.
The limiting point of the process provides an embedding at each node,
which can then be combined with a simpler model such as multinomial logistic regression to make predictions at each node.
This results into a method which is much faster than typical GNNs as the training phase and embedding computation are decoupled,

Remarkably, even though our model is nonlinear, we can still prove a number of theoretical properties about the diffusion process.
In particular, we show that the limiting point of the process is unique and provide a simple, globally convergent iterative algorithm for computing it.
Furthermore, we show that this limiting point is the global optimum of an interpretable  hypergraph SSL loss function defined as the combination of a data fitting term and a Laplacian-like regularizer which aims at reducing a form of ``generalized variance'' on each hyperedge. 
%
Empirically, we find that using the limiting point of our nonlinear hypergraph diffusion as features for a linear model is competitive with state-of-the-art graph and hypergraph neural networks and other diffusions on several real-world datasets.
We also study the effect of the input feature embedding on the classification performance by either removing or modifying the node features. In particular, including the final embedding of hypergraph GNNs as additional features in the diffusion model does not improve accuracy, which provides evidence that our model is sufficient for empirical data.

\section{Problem set-up}
We consider the multi-class semi-supervised classification problem on a hypergraph, in which we are given nodes with features and hyperedges connecting them. A small number of node labels are available and the goal is to assign labels to the remaining nodes. 

Let $H=(V,E)$ be a hypergraph where $V=\{1,\dots,n\}$ is the set of nodes and $E = \{e_1,\dots,e_m\}$ is the set of hyperedges. 
Each hyperedge $e \in E$ has an associated positive weight $w(e) > 0$.
In our setting every node can belong to an arbitrary number of hyperedges.
Let $\delta_i$ denote the (hyper)degree of node $i\in V$, i.e., the weighted number of hyperedges node $i$ participates in, $\delta_i = \sum_{e:i\in e}w(e)$, and let $D\in \RR^{n\times n}$ be the diagonal matrix of the node degrees, i.e., $D = \Diag(\delta_1,\dots,\delta_n)$. Throughout we assume that hypergraph has no isolated nodes, i.e., $\delta_i>0$ for all $i$. This is a standard assumption, as one can always add self-loops or remove isolated vertices.
Let $K$ denote the $n\times m$ incidence matrix of $H$, whose rows correspond to nodes and columns to hyperedges:
\[
K_{i,e} = 
\begin{cases}
1 & i\in e \\
0 & \text{otherwise.}
\end{cases}
\]
To include possible weights on hyperedges, we use a diagonal matrix $W$ defined by 
$
W = \Diag(w(e_1),\dots,w(e_m))
$.

We will represent $d$-dimensional features on nodes in $H$ by a matrix $X\in \RR^{n\times d}$,
where row $x_i = X_{i,:} \in \RR^d$ is the feature vector of $i\in V$. 
Suppose each node belongs to one of $c$ classes, denoted as $\{1,\dots, c\}$, 
and we know the label of a (small) training subset $\mathcal T$ of the nodes $V$.
We denote by  $Y\in \RR^{n\times c}$ the input-labels matrix of the nodes, with rows $y_i$ entrywise defined by $Y_{ij} = (y_i)_j =1$ if node $i$ belongs to class $j$, and $(y_i)_j =0$ otherwise.  
Since we only know the labels for the nodes in $\mathcal T$, all the  $y_i$ for $i\not\in \mathcal T$ are zero, 
while the for $i\in \mathcal T$, $y_i$ has exactly one nonzero entry. (one-hot encoding)

\section{Background and related work}\label{sec:background}

We review basic ideas in hypergraph neural networks (HNNs) for SSL and hypergraph label spreading (HLS),
which will contextualize the method we develop next.

\subsection{Neural network approaches}
Graph neural networks are broadly adopted methods for semi-supervised learning on graphs. 
Several generalizations to hypergraphs have been proposed, and we summarize 
the most fundamental ideas here.

When $\lvert e \rvert=2$ for all $e\in E$, the hypergraph is a standard graph $G$.
The basic formulation of a graph convolutional network (GCN)~\cite{kipf2017} is based on a first-order approximation of
the convolution operator on graph signals \cite{mallat1999wavelet}. 
This approximation boils down to a mapping given by the normalized adjacency matrix of the graph $\bar A=I - L$, where $L$ is the (possibly rescaled) normalized Laplacian 
$L = I -D^{-1/2} A D^{-1/2}$ and  $A$ is the adjacency matrix. 
The forward model for a two-layer GCN is then
$$
Z = \mathrm{softmax}(F) = \mathrm{softmax}(\bar A\sigma(\bar AX \Theta^{(1)})\Theta^{(2)})
$$
where $X\in \RR^{n\times d}$ is the matrix of the   node features, $\Theta^{(1)}$, $\Theta^{(2)}$ are the input-to-hidden and hidden-to-output weight matrices of the network and $\sigma$ is a nonlinear activation function. 
Here, the approximated graph convolutional filter $\bar A$ combines features across nodes that are well connected in the  graph.
For multi-class semi-supervised learning problems, the weights $\Theta^{(i)}$ are then trained minimizing the cross-entropy loss 
over the training set of known labels~$\mathcal T$. 

Several hypergraph variations of this neural network model have been proposed for the more general case $\lvert e \rvert\geq 2$. 
A common strategy is to consider a hypergraph Laplacian $L$ and define an analogous convolutional filter.
One relatively simple approach is to define $L$ as the Laplacian of the clique expanded graph of $H$~\cite{agarwal2006higher,zhou2007hypergraph}, where the hypergraph is mapped to a graph on the same set of nodes by adding a clique among the nodes of each hyperedge.
This is the approach used in HGNN~\cite{feng2019hypergraph},
and other variants use mediators instead of cliques in the hypergraph to perform the graph reduction~\cite{chan2020generalizing}.
HyperGCN~\cite{yadati2019hypergcn} is based on the nonlinear hypergraph Laplacian proposed in~\cite{chan2018spectral,louis2015hypergraph}. 
This model uses a GCN on a reduced graph $G_X = (V, E_X)$ that depends on the features,
where $(u, v) \in E_X$ if and only if $(u,v) = \arg\max_{i,j\in e}\| x_i- x_j\|$, for all hyperedges $e$ of the original hypergraph.
An approximate graph convolutional filter $\bar{\mathcal  A}_{X}$ is then defined in terms of the normalized Laplacian of $G_{X}$ as before. Thus, the two-layer forward model for  HyperGCN is
\begin{align*}
Z =  \mathrm{softmax}(\mathcal  A_{F^{(1)}}\Theta^{(2)}), \quad F^{(1)} = \sigma(\mathcal A_{X} \Theta^{(1)})\, .
\end{align*}

\subsection{Laplacian regularization and label spreading}\label{sec:regularization_methods}
Semi-supervised learning based on Laplacian-like regularization strategies were developed in \cite{zhou2004learning} for graphs and then in \cite{zhou2007hypergraph} for hypergraphs. 
The main idea of these approaches is to obtain a classifier $F$ by minimizing the regularized square loss function
\begin{equation}\label{eq:general-reg-loss}
    \min_F  \ell_{\Omega} := \|F-Y\|^2 + \lambda \,\Omega(F)
\end{equation}
where $\Omega$ is a regularization term that takes into account for the hypergraph structure. 
(Note that only labels --- and not features --- are used here.)
In particular, if $f_i = F_{i,:}$ denotes the $i$-th row of $F$, the clique expansion approach of  \cite{zhou2007hypergraph} defines $\Omega= \Omega_{L^2}$, with
\[
 \Omega_{L^2}(F) = \sum_{e\in E}\sum_{i,j\in e}\frac {w(e)}{|e|} \Big\|\frac{f_i}{\sqrt{\delta_i}} - \frac{f_j}{\sqrt{\delta_j}}\Big\|^2,
\]
while the total variation on hypergraph regularizer proposed in  \cite{hein2013total} is $\Omega=\Omega_{TV}$, where
\[
 \Omega_{TV}(F) = \sum_{e\in E}w(e) \max_{i,j\in e} \|f_i - f_j\|^2
\]
The function $\Omega_{L^2}$ is quadratic, so its gradient directly boils down to the Laplacian of the clique expanded graph of $H$. Thus, HGNN is implicitly applying this regularization. Similarly, the graph construction in HyperGCN is implicitly applying a regularization based on the hypergraph total variation $\Omega_{TV}$.

These two choices of regularizing terms can be solved by means of different strategies. 
As $\Omega_{L^2}$ is quadratic, one can solve \eqref{eq:general-reg-loss} via gradient descent with the learning rate $\alpha = \lambda /(1+\lambda)$ to obtain the simple  method
\begin{equation}\label{eq:HLS}
    F^{(k+1)} = \alpha \bar A_H F^{(k)} + (1-\alpha)Y,
\end{equation}
where $\bar A_H$ is the normalized adjacency matrix of the clique-expanded graph of $H$. The sequence \eqref{eq:HLS} converges to the global solution $F_\star$ of \eqref{eq:general-reg-loss} for any starting point and the limit $F_\star$ is entrywise nonnegative. This method is usually referred to as Hypergraph Label Spreading (HLS) as the iteration in \eqref{eq:HLS} takes the initial labels $Y$ and ``spreads'' or ``diffuses'' them throughout the vertices of the  hypergraph $H$, following the edge structure of its clique-expanded graph. 
%
%
%
%

The total variation inspired regularizer $\Omega_{TV}$ is related to the $1$-Laplacian energy~\cite{buhler2009spectral,tudisco2018nodal} and
has advantages for hyperedge cut interpretations. 
However, although $\Omega_{TV}$ is convex, it is not differentiable, and computing \eqref{eq:general-reg-loss} requires more complex and computationally demanding methods~\cite{hein2013total,zhang2017re}.
Unlike HLS in \eqref{eq:HLS}, this is not easily interpreted as a label diffusion. 

\section{Hyperedge variance regularization and nonlinear diffusion}\label{sec:nonlinear-HLS}

In this section, we propose a new hypergraph regularization term $\Omega_\mu$ which, rather than minimizing the distance between each node pair on a hyperedge,  aims at reducing the variance (or a generalized variance)  across the hyperedge nodes. Precisely,  consider the regularization term of the form
\begin{equation}\label{eq:omega_star}
 \Omega_\mu(F) = \sum_{e\in E}\sum_{i\in e}w(e)\Big\|\frac{f_i}{\sqrt{\delta_i}}-\mu\Big(\Big\{\frac{f_j}{\sqrt{\delta_j}} : j \in e\Big\}\Big)\Big\|^2
 \end{equation}
where $\mu(\cdot)$ is a function that combines node embeddings on each hyperedge. When $\mu$ is the mean $\mu(\{z_j : j\in e\}) = \frac 1 {|e|}\sum_{j\in e}z_j$, 
the right hand side in \eqref{eq:omega_star} coincides exactly with the variance of $f_i/\sqrt{\delta_i}$ on the hyperedge~$e$. 

Note that, if $F$ is the matrix with rows $F_{i,:}=f_i$, the  mean of $f_i/\sqrt{\delta_i}$ over the hyperedge $e$ can be written as the $e$-th row of the matrix $D_E^{-1}K^\top D^{-1/2}F$, where $D_E$ denotes the $m\times m$ diagonal matrix with diagonal entries $D_{e,e} = |e|$.
 Here we use this observation to define a family of functions $\mu$ that generalizes the mean. Precisely, let 
\begin{equation}\label{eq:mu_e}
\mu_{\sigma,\varrho}\Big(\Big\{\frac{f_j}{\sqrt{\delta_j}} : j \in e\Big\}\Big) = \sigma(K^\top \varrho(D^{-1/2}F))_{e,:} 
\end{equation}
where   $\sigma$ and $\varrho$ are (in general, nonlinear) operators that describe the way an embedding $F$ is transformed and combined across the hyperedges.
For example,  when $\sigma$ and $\varrho$ are diagonal operators (similar to activation functions) --- i.e.\ they are such that $\sigma(F)_{ij} = \sigma_i(F_{ij})$ for some functions $\sigma_1, \sigma_2,\dots:\RR\to \RR$ (and the same for $\varrho$) --- we have that 
\[
\mu_{\sigma,\varrho}\Big(\Big\{\frac{f_j}{\sqrt{\delta_j}} : j \in e\Big\}\Big) =  \sigma_e\Big(\sum_{j \in e} \varrho_j \big(\frac {f_j} {\sqrt{\delta_i}} \big)\Big).
\]
An important example of $\sigma$ and $\varrho$  of this form, which we will use in all our experiments, is
\begin{equation}\label{eq:choice_of_sigma_rho}
     \varrho(Z_1) := Z_1^p, \qquad  \sigma(Z_2) :=    (D_E^{-1}Z_2)^{1/p},
\end{equation}
where the powers are taken entry-wise and $Z_i$ are matrices of appropriate size. With these choices, for every $e \in E$  we have that $\mu_{\sigma, \varrho}(\{f_i/\sqrt{\delta_i},i\in e\}) = \mathrm{mean}_p\{f_i/\sqrt{\delta_i} : i\in e\}$ is the $p$-power mean of the normalized feature vectors $f_i/\sqrt{\delta_i}$ of the nodes $i$ in the hyperedge $e$, where
\[
\textstyle {\mathrm{mean}_p\{z_i : i\in e\} = \big( \frac 1 {|e|} \sum_{i\in e} z_i^p\big)^{1/p} \, .}
\]
With this choice of $\sigma$ and $\varrho$  the regularization term \eqref{eq:omega_star} reads 
\begin{equation}\label{eq:p-mean-Omega}
    \Omega_{\mu_{\sigma, \varrho}}(F) = \sum_{e\in E}\sum_{i\in e}w(e)\Big\|\frac{f_i}{\sqrt{\delta_i}}-\mathrm{mean}_p\Big\{\frac{f_j}{\sqrt{\delta_j}} : j \in e\Big\}\Big\|^2 
\end{equation}
that is, the embedding $F$ that minimizes $\Omega_{\mu_{\sigma, \varrho}}$ minimizes the variation of each node embedding $f_i$ from the $p$-power mean of the embeddings of  the nodes in each hyperedge $i$ participates to. In particular, when $p=2$, the regularization term \eqref{eq:p-mean-Omega} computes the variance of all the nodes in each of the hyperedges.
Note that, for nonnegative embeddings, other special cases of $\mathrm{mean}_p$ include the geometric mean (for $p\to 0$), the harmonic mean (for $p=-1$) as well as the minimum and maximum functions  (for $p\to -\infty$ and $p\to +\infty$, respectively).

\subsection{Nonlinear diffusion method}
Consider the regularized loss function $\ell_{\Omega}$ in \eqref{eq:general-reg-loss} with $\Omega = \Omega_{\mu_{\sigma,\varrho}}$. Unlike the  clique-expansion case, $\Omega = \Omega_{L^2}$, $\ell_{\Omega_{\mu_{\sigma,\varrho}}}$ is non-quadratic and non-convex in general. Despite this fact, we will show below that the global solution to $\min_F \ell_{\Omega_{\mu_{\sigma,\varrho}}}(F)$ can be computed via a simple hypergraph diffusion algorithm similar to a nonlinear version of HLS \eqref{eq:HLS}, provided we restrict to the set of embeddings with nonnegative entries.  We introduce the diffusion model next.

Recall that in our setting each node $i\in V$ has a label-encoding vector $y_i$  ($y_i$ is the all-zero vector for the initially unlabeled points $i\not\in \mathcal T$) and a feature vector $x_i$.
Thus, each node in the hypergraph has an initial $(c+d)$-dimensional embedding, which forms an input matrix 
$U = [Y\,\; X]$. 

The proposed hypergraph semi-supervised classifier uses the normalized limit point of the nonlinear diffusion process 
\begin{equation}\label{eq:nonlinear_HLS}
     F^{(k+1)} = \alpha \, \Phi (F^{(k)}) + (1-\alpha) \, U  \, .
\end{equation}
where the  nonlinear diffusion map $\Phi$ is a form of nonlinear Laplacian operator defined as 
\begin{equation}\label{eq:diffusion_map}
    \Phi(F) = D^{-1/2}KW\sigma(K^\top \varrho(D^{-1/2}F))\, .
\end{equation}

The limit point of the diffusion process  \eqref{eq:nonlinear_HLS}  results in a new embedding 
$F_\star = [Y_\star\, X_\star] \in \RR^{n\times(c+d)}$. We will show in \S\ref{sec:convergence} that such limit exists, is unique and minimizes a normalized version of the SSL regularized loss $\ell_{\Omega_{\mu_{\sigma,\varrho}}}$. We will then use $F_\star$ to train a logistic multi-class classifier 
based on the known labels $i \in \mathcal T$ and their new embedding $F_\star$. 
Thus, unlike GNNs, the training phase and the computation of the embedding $F_\star$ are decoupled, and thus are much faster. The overall SSL algorithm is detailed in Algorithm~\ref{alg:HLS}. Note that the convergence of \eqref{eq:nonlinear_HLS} is not trivial, due to the nonlinearity of $\Phi$. We further comment on this issue in the appendix.

\begin{algorithm}[t]
  \caption{\algname{}: Nonlinear Hypergraph Diffusion}\label{alg:HLS}
  \begin{algorithmic} 
     \STATE {\bfseries Input:}\\
        $\bullet$ Incidence matrix $K$ and weights matrix $W$; \\ 
        $\bullet$ mixing mappings $\sigma, \varrho$ as in \eqref{eq:choice_of_sigma_rho};\\
        $\bullet$ normalization mapping $\varphi$ as in \eqref{eq:varphi};\\
        $\bullet$ label $Y\in \{0,1\}^{n \times c}$ and feature $X\in \RR^{n\times d}$ matrices; \\
        $\bullet$ regularization coefficient $\alpha \in (0,1)$ and stopping tolerance $\textnormal{tol}$.\\[0.25em]
        \STATE Shift and scale $U$ to obtain $U>0$, e.g., via \eqref{eq:label-smoothing} when  $X\geq 0$
         \STATE $U \gets U/\varphi(U)$ 
         \STATE $F^{(0)} \gets U$
        \REPEAT 
            \STATE $G \gets  \alpha \Phi(F^{(k)}) + (1-\alpha) U$ 
            \STATE $F^{(k+1)} \gets G / \varphi(G)$ 
        \UNTIL{$\|F^{(k+1)}-F^{(k)}\|/\|F^{(k+1)}\| < \textnormal{tol}$}
        \STATE  New node embedding: $F_\star = [Y_\star\,\, X_\star] \gets F^{(k+1)}$
        \STATE Optimize $\Theta$ for $Z_\star\gets \text{softmax}(F_\star\Theta)$ to minimize cross-entropy. \\[.25em]
        \STATE \textbf{Output:}\\ Prediction $\arg\max_{c} (Z_\star)_{i,c}$ for class of unlabeled node~$i$. \\
  \end{algorithmic}
\end{algorithm}

\subsection{Related nonlinear diffusion models}
Our nonlinear diffusion process \eqref{eq:nonlinear_HLS}  propagates both input node label and feature embeddings through the hypergraph in a manner similar to \eqref{eq:HLS}, but allowing for nonlinear activations, which increases modeling power.
Firstly, $\Phi$ is a generalization of the normalized adjacency matrix of the clique-expansion hypergraph  $\bar A_H$ \eqref{eq:HLS}. 
Secondly, for  a standard graph, i.e., a hypergraph where all the edges have exactly two nodes,
$KWK^\top = A + D$ where $A$ is the adjacency matrix of the graph and $D$ is the diagonal matrix of the weighted node degrees.
Similarly, for a general hypergraph $H$, we have the identity $KWK^\top = A_H+D$, where $A_H$ is the adjacency matrix of the clique-expanded graph   associated with $H$. 
Then, when $\sigma = \varrho = \mathrm{id}$, $\Phi$ coincides with 
\begin{equation}\label{eq:norm_adj_hypergraph}
    D^{-1/2}KWK^\top D^{-1/2} = \bar A_H + I\, ,
\end{equation}
the  normalized adjacency matrix of the clique-expansion hypergraph \cite{zhou2007hypergraph,feng2019hypergraph} as in \eqref{eq:HLS}.

When $\sigma$ and $\varrho$ are not linear, $\Phi$ can represent a broad family of nonlinear diffusion mappings, with special cases used in different context. For example, in the graph case, if  $\varrho=\mathrm{id}$ and $\sigma(x) = |x|^{p-1}\mathrm{sign}(x)$, then $\Phi$ boils down to the graph $p$-Laplacian operator \cite{saito2018hypergraph,elmoataz2008nonlocal,buhler2009spectral}. Exponential and logarithmic based choices of $\sigma$ and $\varrho$ give rise to nonlinear Laplacians used to model  chemical reactions \cite{rao2013graph,van2016network} as well as to model consensus dynamics and opinion formation in hypergraphs \cite{neuhauser2021consensus}. Trigonometric functions such as $\sigma(x) = \sin(x)$ are used to model graph and hypergraph oscillators  \cite{schaub2016graph,battiston2021physics,millan2020explosive}. In the context of semi-supervised learning, nonlinear diffusion mappings based on entrywise powers and generalized means are used, for example in \cite{ibrahim2019nonlinear,tudisco2021nonlinear}. Similarly to our proposed operator, the diffusion map of \cite{tudisco2021nonlinear} generalizes the classical linear label spreading by combining labels on hyperedges via $p$-power means. However, the proposed approach only considers node labels and is based on a 3-rd order tensor specifically designed for $3$-uniform hypergraphs where the hyperedges are cliques of order 3 (triangles), obtained from an input graph data. As tensor-based approaches are limited to uniform hypergraphs, applying that method to general hypergraphs would require to split the hyperedges into batches of same sizes and compute the corresponding adjacency tensors. Compared to our proposed incidence matrix model, this is computationally significantly more demanding both because it requires the computation of several high order tensors and because the multiplication operations with tensors are more expensive than those with matrices (which use  fast BLAS routines from, e.g., NumPy).

\subsection{Convergence}\label{sec:convergence}
The convergence of \eqref{eq:nonlinear_HLS} is discussed in  Theorem \ref{thm:main} below, where we show that, under mild assumptions on $\sigma$ and $\varrho$, \eqref{eq:nonlinear_HLS} always converges, provided the embedding is suitably normalized at each step. Moreover, the result shows that the nonlinear hypergraph filter $\Phi$ eventually generates an embedding that minimizes a regularized loss function of the form \eqref{eq:general-reg-loss}, with regularization term  $\Omega=\Omega_{\mu_{\sigma,\varrho}}$. The proof of Theorem \ref{thm:main} is moved to the appendix.

In the following, we write $F\geq 0$ (resp.\ $F>0$) to indicate that $F$ has nonnegative (resp.\ positive) entries. Moreover,  we write $\sigma\in \hom_+(a)$ to denote that  $\sigma$ is positive 
and homogeneous of degree $a$, i.e. that the following holds for $\sigma$: (1) $\sigma(F) >0$ for all $F>0$; 
and  (2) $\sigma(\lambda F) = \lambda^a \sigma(F)$ for all $\lambda>0$ and all $F>0$.

Note that the class of  operators $\hom_+(a)$ is quite general and it includes, for example different forms of LeakyReLU functions $\sigma(F) = \max\{0,F^a\}\pm \varepsilon \max\{0,-F^a\}$  as well as the family of homogeneous nonnegative generalized polynomial (polynomial with real powers) operators, defined as 
$$
\sigma(F)_{i,:} = \sum_{j=1}^n B^{(i,j)} f_1^{\alpha_1^{(i,j)}}\cdots f_n^{\alpha_n^{(i,j)}}
$$
for any nonnegative coefficients $\alpha_j^{(i,j)}$ and any nonnegative matrices $B^{(i,j)}$, as long as  $\sum\alpha_1^{(i,j)}+\cdots +\alpha_n^{(i,j)}=a$, i.e. the sum of the powers is constant for all $i,j$, to ensure $\sigma(\lambda F) = \lambda^a \sigma(F)$ for all $\lambda>0$. 

\renewcommand{\arraystretch}{1}
\begin{table*}[t!]
\caption{Details for the real-world hypergraph datasets used in the experiments. }
\label{tab:datasets}
\centering
{
\resizebox{16cm}{!}{%
    \begin{tabular}{c l c c c c c c}
    \toprule
               & &\bf DBLP & \bf Pubmed & \bf Cora & \bf Cora & \bf Citeseer & \bf Foodweb \\
    ~    & ~            & co-authorship & co-citation & co-authorship & co-citation & co-citation & carbon-exchange \\ 
    \midrule
    $|V|$ & ($\#$nodes) & 43413 & 19717 & 2708 & 2708 & 3312 & 122 \\
    $|E|$ & ($\#$hyperedges) & 22535 & 7963 & 1072 & 1579 & 1079 & 141233 \\
    $d$ & ($\#$features) & 1425 & 500 & 1433 & 1433 & 3703 & 0\\
    $c$ & ($\#$labels) & 6 & 3 & 7 & 7 & 6 & 3 \\
    \bottomrule
    \end{tabular}
}
}
\end{table*}

\begin{theorem}\label{thm:main}
Let $\Phi$ be defined as in \eqref{eq:diffusion_map} and  let $\mu_{\sigma, \varrho}$ be defined as in \eqref{eq:mu_e}. Let $f_i=F_{i,:}$ denote the $i$-th row of $F$ and define the real-valued function 
\begin{equation}
    \varphi(F) = 2 \sqrt{\sum_{e\in E}  w(e) \Big\|\mu_{\sigma,\varrho}\Big(\Big\{\frac{f_j}{\sqrt{\delta_j}}, j \in e\Big\}\Big)\Big\|^2} \, . \label{eq:varphi}
\end{equation}
Assume that $\sigma \in \hom_+(a)$ and $\rho\in \hom_+(b)$ for some $a,b\in \RR$.  Let $U$ be an entrywise positive input embedding and  let $\alpha \in [0,1]$. 
If $ab = 1$ and if $\Phi$ is differentiable and such that $\Phi(F)\geq \Phi(\tilde F)$ for all $F\geq \tilde F>0$, then for any starting point $F^{(0)}\geq 0$,  the sequence 
\begin{equation}\label{eq:general_feature_spreading}
\begin{cases}
\tilde F^{(k)} = \alpha \, \Phi (F^{(k)}) + (1-\alpha) \, U  & \\
F^{(k+1)} = \tilde F^{(k)} / \varphi(\tilde F^{(k)}) & 
\end{cases}
\end{equation}
converges to a unique fixed point $F_\star$  of \eqref{eq:nonlinear_HLS}, such that   $\varphi(F_\star) = 1$, $F_\star>0$. 
Moreover,  $F_\star$  is the solution~of
\[
\begin{cases}
\displaystyle{\min_{F}}\,\,\,\, \Big\|F-\frac U {\varphi(U)}\Big\|^2 + \lambda \, \Omega_{\mu_{\sigma,\varrho}}(F) &\\  
\mathrm{subject}\,\,\mathrm{to}\,\,\, F\geq 0,\, \,  \varphi(F)=1, & 
\end{cases}
\]
where  $\lambda = \alpha/(1-\alpha)$.
\end{theorem}

Note that the choices of $\sigma$ and $\varrho$ in \eqref{eq:choice_of_sigma_rho}, which lead to the $p$-power mean regularization term \eqref{eq:p-mean-Omega}, and the corresponding diffusion operator $\Phi$ satisfy all the assumptions of Thm~\ref{thm:main}. \!\!\!

\subsection{Algorithm details and limitations}\label{sec:limit}

Once the new node embedding $F_\star$ is computed via \eqref{eq:general_feature_spreading}, we use it to infer the labels of the non-labeled data points via a simple softmax output layer which minimizes  cross-entropy (see Algorithm \ref{alg:HLS}). 

Similar to  HLS, the parameter $\alpha$ in Alg.~\ref{alg:HLS} yields a convex combination of the diffusion mapping $\Phi$ and the ``bias'' $U$, allowing to tune the contribution given by the homophily along the hyperedges and the one provided by the input features and labels. In other words, in view of Theorem \ref{thm:main},  $\alpha$ quantifies the strength of the regularization parameter $\lambda$, 
which allows us to tune the  contribution of the regularization term $\Omega_{\mu_{\sigma,\varrho}}$ over the data-fitting term $\|F-U/\varphi(U)\|$.

A requirement for our main theoretical results  is entrywise positivity of the input embedding $U$. While this is a limitation of the theory and the methods, it turns out to not be that stringent in practice.
If $X\geq 0$, i.e.\ we have nonnegative node features, we can easily obtain a positive embedding by performing an initial label smoothing \cite{muller2019does,szegedy2016rethinking}, i.e.\ we choose a small $\varepsilon>0$ 
and set  
\begin{equation}\label{eq:label-smoothing}
    U_\ep=(1-\varepsilon)[Y\,\, X] + \varepsilon \one \one\, ,
\end{equation}
being $\one\one$ the all-one matrix of the appropriate size. Note that nonnegative input features $X\geq 0$ are not uncommon. For instance, bag-of-words, one-hot encodings, and binary features in general are all nonnegative. 
In fact, for all of the real-world datasets we consider in our experiments, the features are nonnegative.
Similarly, if some of the input features have negative values (e.g., features coming from a word embedding), one could perform other preprocessing (e.g., shift on all embeddings) to get the required~$[Y   X]>\!0$.

Another implicit requirement of the proposed method is the assumption that hypergraph datasets at hand are homophilic. However, this limitation is only marginal as homophily is common in empirical data and is indeed a key principle behind many graph and hypergraph methods.

\section{Experiments}\label{sec:experiments}

\begin{table*}[t]
\newcommand{\smallpm}[1]{{\footnotesize{$\pm$#1}}}
\def\first#1{\setlength{\fboxsep}{2pt}\fcolorbox{black}{black!15}{\textbf{#1}}}
\def\second#1{{\setlength{\fboxsep}{2pt}\fcolorbox{black!50}{black!5}{#1}}}
\renewcommand{\arraystretch}{1}

\caption{Accuracy (mean $\pm$ standard deviation) over five random samples
         of the training nodes $\mathcal T$.  We compare \algname{} 
         and the six baseline methods (APPNP, HGNN, HyperGCN, SGC, SCE, HTV).
         Overall, 
         \algname{} is more accurate than the baselines.}
\label{tab:realworld}
\centering
{
\resizebox{\textwidth}{!}{%
\begin{tabular}{l  c   ccccccc}
\toprule
& \bf Method &\bf \algname{} & \bf APPNP & \bf HGNN & \bf HyperGCN & \bf SGC & \bf SCE & \bf HTV \\
\midrule
 Data & \% labeled &   &   &   &   &   &  &  \\
\midrule
\bf Citeseer & 4.2\%  & \textbf{72.13} \smallpm{1.00} & 63.51 \smallpm{1.39} & 61.78 \smallpm{3.46} & 50.94 \smallpm{8.27} & 52.66 \smallpm{2.18} & 61.28 \smallpm{1.61} & 29.63\smallpm{0.3}\\ 
\bf Cora-author & 5.2\%  & \textbf{77.33} \smallpm{1.51} & 71.34 \smallpm{1.60} & 63.11 \smallpm{2.73} & 61.27 \smallpm{1.06} & 30.46 \smallpm{0.22} & 71.96 \smallpm{2.18} & 44.55\smallpm{0.6}\\ 
\bf Cora-cit & 5.2\%  & \textbf{83.13} \smallpm{1.11} & 82.08 \smallpm{1.61} & 62.88 \smallpm{2.26} & 62.78 \smallpm{2.73} & 29.08 \smallpm{0.25} & 79.85 \smallpm{1.91} & 35.60\smallpm{0.8} \\ 
\bf DBLP & 4.0\%  & \textbf{89.63} \smallpm{0.12} & 88.94 \smallpm{0.07} & 73.82 \smallpm{0.71} & 70.02 \smallpm{0.10} & 43.61 \smallpm{0.17} & 87.50 \smallpm{0.19} & 45.19\smallpm{0.9} \\ 
\bf Foodweb & 5.0\%  & 64.09 \smallpm{5.94} & \textbf{69.12} \smallpm{3.30} & 57.09 \smallpm{2.33} & 56.14 \smallpm{3.85} & 57.45 \smallpm{0.47} & 63.50 \smallpm{4.78} & 57.23\smallpm{0.9} \\ 
\bf Pubmed & 0.8\%  & \textbf{82.81} \smallpm{2.16} & 81.50 \smallpm{1.18} & 72.57 \smallpm{1.03} & 78.11 \smallpm{0.99} & 54.30 \smallpm{1.11} & 77.57 \smallpm{2.34} & 47.04\smallpm{0.8}\\ 

\bottomrule
\end{tabular}%
}
}
\end{table*}

We now evaluate our method on several real-world hypergraph datasets (Table~\ref{tab:datasets}).
We use five co-citation and co-authorship hypergraphs: Cora co-authorship, Cora co-citation, Citeseer, Pubmed \cite{sen2008collective} and DBLP \cite{rossi2015network}. All nodes in the  datasets are documents, features are given by the content of the abstract and  hyperedge connections are based on either co-citation or co-authorship. The task for each dataset is  to  predict  the  topic  to  which  a  document  belongs. 
We also consider a foodweb hypergraph, where the nodes are organisms and hyperedges represent directed carbon exchange in the Florida bay \cite{foodweb}. Here we predict the role of the nodes in the food chain. This hypergraph has no features, so only labels are used for \algname{} while we use a one-hot encoding for the baselines.

\begin{figure*}[!t]
    \centering
    \includegraphics[width=\textwidth,clip,trim=0cm .4cm 0cm .4cm]{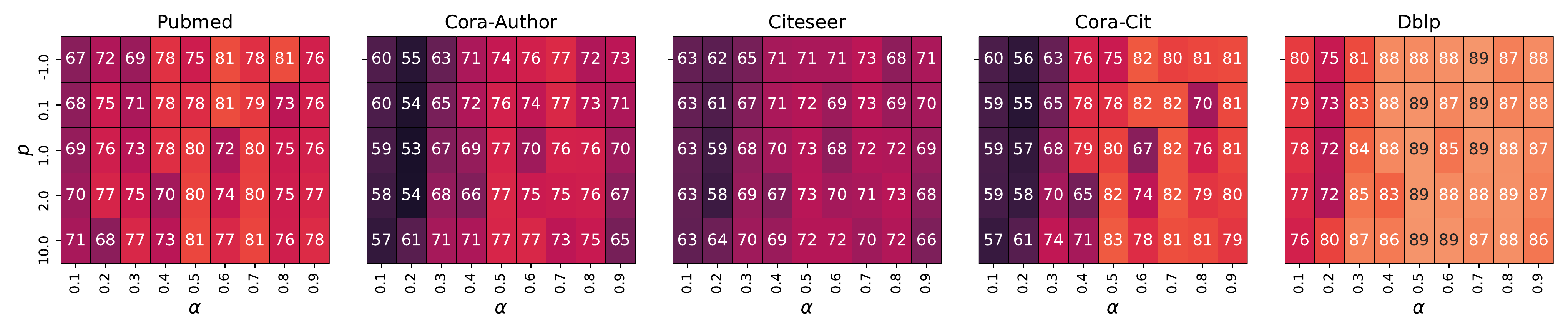}
    \caption{Performance of the proposed  \algname{} for varying $p$ and $\alpha$ parameters.\vspace{-1.5em}}
    \label{fig:varying_alpha}
\end{figure*}

We compare our method to six baselines. 
Two are hypergraph convolutional network, designed to work specifically for hypergraphs:
\begin{itemize}[topsep=0pt,noitemsep,leftmargin=*]
  
    \item \textbf{HGNN}~\cite{feng2019hypergraph} This is a hypergraph neural network model that uses the clique-expansion Laplacian  \cite{zhou2007hypergraph,agarwal2006higher} for the hypergraph convolutional filter.
    
    \item \textbf{HyperGCN}~\cite{yadati2019hypergcn} This is a hypergraph convolutional network model
    with regularization similar to the total variation (see also \S\ref{sec:background}). 
    There are three architectural variants (1-HyperGCN, FastHyperGCN, HyperGCN), and
    we report whichever gives the best performance.
\end{itemize}
One is a hypergraph Laplacian regularization method:
\begin{itemize}[topsep=0pt,noitemsep,leftmargin=*]
    \item \textbf{HTV} \cite{zhang2017re} This is a confidence-interval subgradient-based method that minimizes the $1$-Laplacian inspired  loss \eqref{eq:general-reg-loss} with $\Omega=\Omega_{TV}$. In our experiments, this method outperformed other
    label spreading and Laplacian regularization techniques such as the original PDHG strategy of \citet{hein2013total}, the HLS method~\cite{zhou2007hypergraph} and the $p$-Laplacian approach of~\citet{saito2018hypergraph}.
\end{itemize}
The remaining three baselines, instead, are graph convolutional networks designed for graph data, which we apply to the clique-expanded version of the considered hypergraphs:  
\begin{itemize}[topsep=0pt,noitemsep,leftmargin=*]
    \item \textbf{APPNP}~\cite{APPNP} This is a graph convolutional network model combined with PageRank. The authors of this paper introduce a personalized propagation of neural predictions and its approximation based on the relationship between GCN and PageRank.
    
    \item \textbf{SGC}~\cite{SGC} This is a graph convolutional network model~ without nonlinearities.
    
    \item \textbf{SCE}~\cite{SCE} This is a graph convolutional network model inspired by a sparsest-cut problem, where unsupervised network embedding is learned only using negative samples for training.
\end{itemize}


Table \ref{tab:realworld} shows the size of the training dataset for each network and compares the accuracy (mean $\pm$ standard deviation) of \algname{} against the different baselines. 
For each dataset, we use five trials with different samples of the training nodes $\mathcal T$.
All of the algorithms that we use have hyperparameters. For the baselines we use the reported tuned hyperparameters. 
For all of the neural network-based models, we use two layers and 200 training epochs, 
following \cite{yadati2019hypergcn} and \cite{feng2019hypergraph}. 
For our method and HTV, which have no training phase, we run 5-fold cross validation with label-balanced 50/50 splits to choose $\alpha$ from $\{0.1, 0.2, \ldots, 0.9\}$ and $p$ from $\{1, 2, 3, 5, 10\}$.   Precisely, we split the data into labeled and unlabeled points. We split the labeled points into training and validation sets of equal size (label-balanced 50/50 splits) and we choose the parameters based on the average performance on the validation set over 5 random repeats. Then, we assess the performance on the held out test set, which is comprised of all the initially non-labeled points. We repeat this process 5 times and we choose the value of $\alpha,p$ that gives the best mean accuracy. These values differ across the different random samplings of the training dataset. Thus,  in order to highlight how the performance is affected by different values of $\alpha$ and $p$, we show in Figure~\ref{fig:varying_alpha} the mean accuracy over 10 runs of \algname{}, for all of the considered values of $\alpha$  and $p$. 

All the datasets we use here have nonnegative input embedding $[Y\ X]$ which we preprocess via label smoothing as in \eqref{eq:label-smoothing}, with $\ep = \mathtt{1e-6}$. Our experiments have shown that different choices of $\ep$ do not influence the classification performance of the algorithm.

Due to its simple regularization interpretation, we choose $\sigma$ and $\varrho$ to be the $p$-power mean considered in \eqref{eq:choice_of_sigma_rho}, for various $p$. 
When varying $p$, we change the nonlinear activation functions that define the final embedding $F_\star$.
%
%
%
%
%
Our proposed nonlinear diffusion method performs overall very well. Interestingly, the best competitors are not hypergraph-oriented methods but rather graph methods directly applied to the clique-expanded graph. In particular, APPNP is the strongest baseline, and this method also strongly relies on diffusions. Moreover, the poorer performance observed for food web dataset (which has no features) highlights the ability of \algname{} to create meaningful feature embeddings, in addition to propagating labels.


\begin{figure}[!ht]
\centering
    \includegraphics[width=.7\columnwidth, clip,trim=0 0cm 0 0cm]{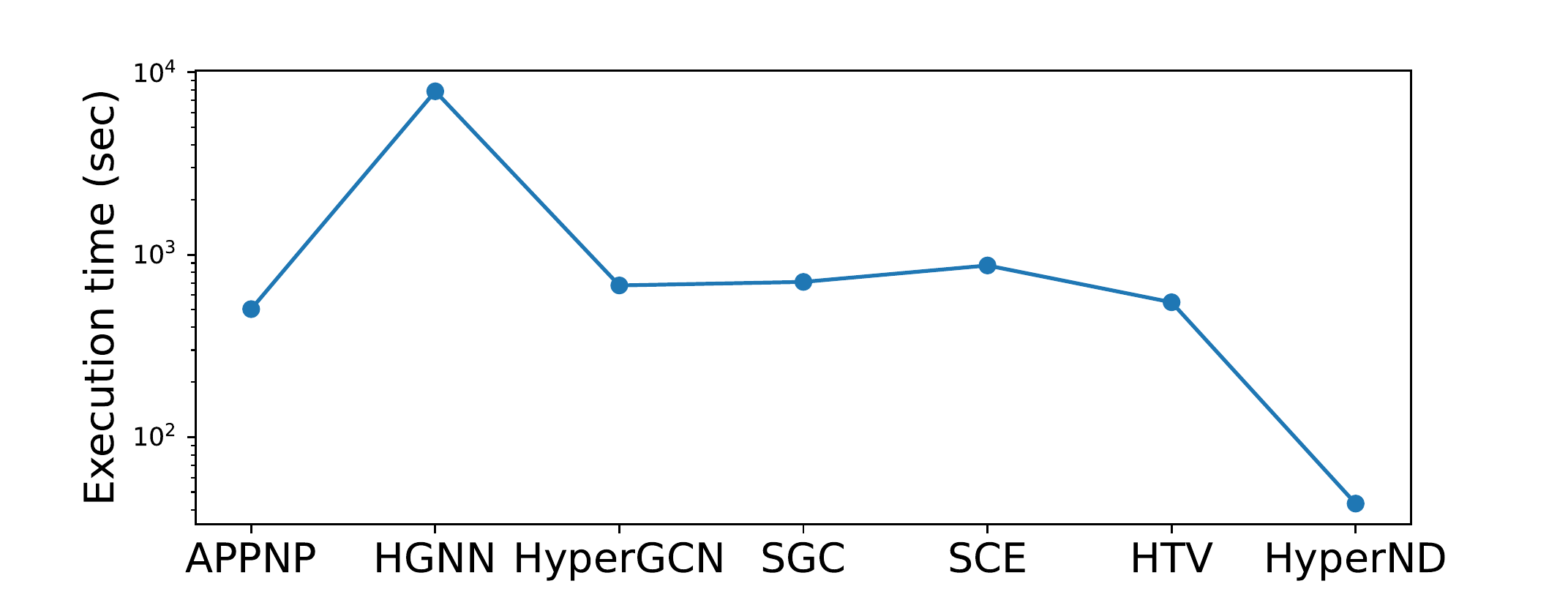}
    \caption{Execution time on the largest dataset DBLP (for one hyper-parameter setting in each case). 
    All methods are comparable on small datasets. \vspace{-1.1em}
}
\label{fig:time}
\end{figure}

We also point out that since \algname{} is a  forward model, it can be implemented efficiently.
The cost of each iteration of  \eqref{eq:HLS}  is dominated by the cost of the two matrix-vector products with the matrices $K$ and $K^\top$,
both of  which only require a single pass over the input data and can be parallelized with standard techniques. 
Therefore, \algname{} scales linearly with the number and size of  the hyperedges, i.e.\ the size of the data. 
Thus, it is typically cheap to compute (similar to standard hypergraph LS) and is overall faster to train than a two-layer GCN.
Training times are reported in Figure \ref{fig:time}, where we compare mean execution time over ten runs for all the methods on  DBLP. The execution times are  similar on other datasets. 
For \algname{}, we show mean execution time over the five random choices of $p$. \algname{} is up to 2 order of magnitudes faster than the baselines.

To further highlight the learning capabilities of the diffusion map we are proposing, we  present one more test below. 
The diffusion map \eqref{eq:HLS} generates a new embedding $F_\star=[Y_\star\,\, X_\star]$ from the fixed point of a purely forward model. 
This model yields a new feature-based representation $X_\star$, similar to the last-layer embedding of any neural network approach.
A natural question is whether or not $X_\star$ is actually a better embedding. 
To this end, in the next experiment we consider four node embeddings $\bar F$ and train a classifier via cross-entropy minimization of $Z = \text{softmax}(\bar F\Theta)$, optimizing $\Theta$.
Specifically, we consider the following:
\begin{enumerate}[topsep=0pt,noitemsep,leftmargin=20pt]
    \item[(E1)] $\bar F= Y_\star$. We run a nonlinear ``purely label'' spreading iteration, by setting $U=Y$ in \eqref{eq:nonlinear_HLS}.
     By Theorem \ref{thm:main}, this embedding is a Laplacian regularization method. 
    \item[(E2)] $\bar F=[Y_\star\,\, X_{\mathrm{hgcn}}]$, where $X_{\mathrm{hgcn}}$ is the embedding generated by HyperGCN before the softmax.
    \item[(E3)] $\bar F=F_\star=[Y_\star\,\, X_\star]$,  the limit point \eqref{eq:nonlinear_HLS} of our \algname{}. This is the embedding used for the results in Table \ref{tab:realworld} and Figure \ref{fig:varying_alpha}.
    \item[(E4)] $\bar F=[Y_\star\,\, X_\star\,\, X_{\mathrm{hgcn}}]$.  This combines the representations of our \algname{}  and HyperGCN.
\end{enumerate}

\begin{figure*}[t!]
    \centering
    \input{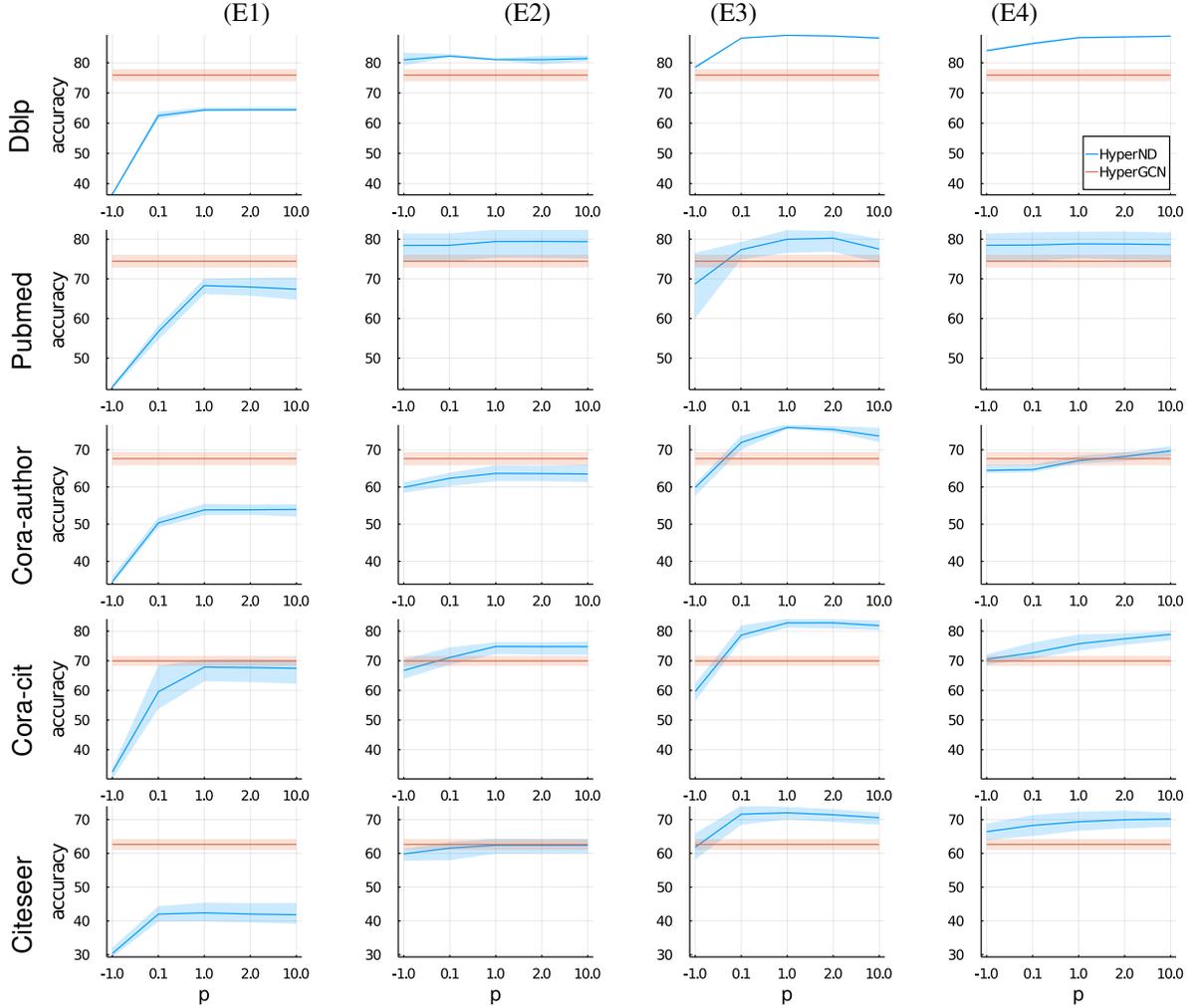} 
    \caption{Accuracy (mean and standard deviation) of multinomial logistic regression classifier, using different combinations of features obtained from embeddings (E1)--(E4).\vspace{-1em}}
    \label{fig:embeddings_comparison}
\end{figure*}

Figure \ref{fig:embeddings_comparison} shows the accuracy for these embeddings with various values of $p$ for the $p$-power mean in \algname{}. The best performance is obtained by the two embeddings that contain our learned features $X_\star$: (E3) and (E4). In particular, while (E4) includes the final-layer embedding of HyperGCN, it does not improve accuracy over (E3). 

\appendix

\section{On the convergence of the nonlinear diffusion.}\label{ap:on_the_convergence}

\begin{figure}[h]
\centering
\includegraphics[width=.5\textwidth]{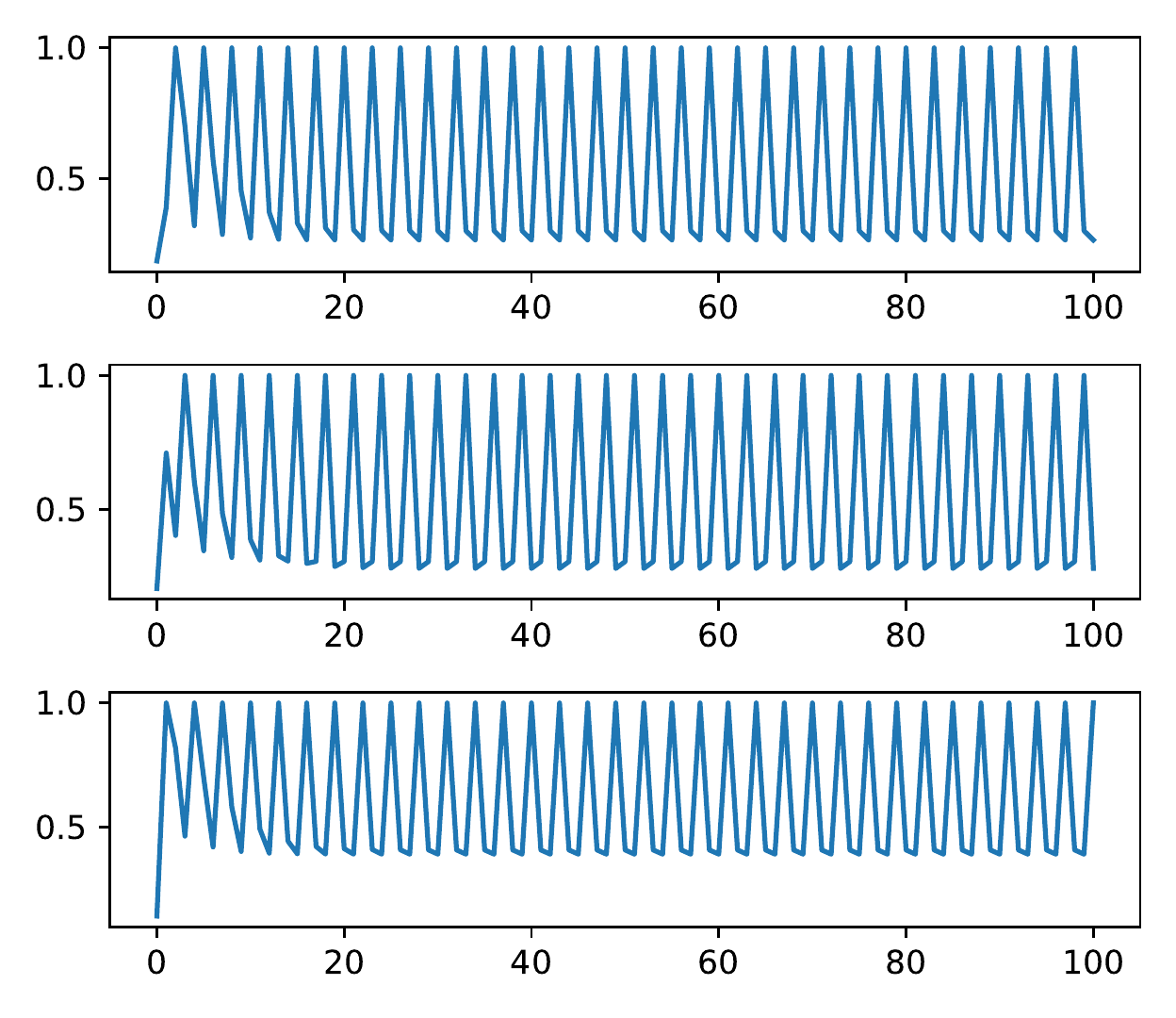}
\caption{Example of a nonconvergent nonlinear diffusion. Each panel shows the behaviour of one of the three coordinates of $x_k$ in \eqref{eq:example} as a function of $k$, for $k=1,\dots,100$.}
\label{fig:no-convergence}
\end{figure}

Our classification method is based on the nonlinear diffusion process on the hypergraph  described in \eqref{eq:nonlinear_HLS} and \eqref{eq:diffusion_map}. 
Unlike the linear case, where the long term behaviour of the discrete dynamical systems can be easily studied, when linear mappings are combined with nonlinearities, neither the existence nor the  uniqueness of a limit point  is obvious for the nonlinear discrete diffusion model \eqref{eq:nonlinear_HLS}. This makes the convergence result in Theorem \ref{thm:main} particularly remarkable.  In fact, already in the vector case, it is easy to find non convergent normalized iterates of the type $\tilde x_{k+1}=  \Phi(x_k)+y$, $x_{k+1}=\tilde x_{k+1}/\|\tilde x_{k+1}\|$, or iterates with multiple fixed points. 
Consider, for  example the two iterations
\begin{equation}\label{eq:example}
z_{k+1} = A(Az_k)^{1.5} + y\, , \qquad \text{and} \qquad \begin{cases}
\tilde x_{k+1} = A(Ax_k)^{1.5} + y\\
x_{k+1} = \tilde x_{k+1} / \|\tilde x_{k+1}\|_\infty
\end{cases}
\end{equation}
where the power is taken component-wise and where $A$ and $y$ are 3-dimensional and chosen as follows:
$$
\textstyle{A = \begin{bmatrix}0 & 1 & 0 \\ 0 & 0 & 1 \\ 1 & 0 & 0 \end{bmatrix}, \qquad  y = \begin{bmatrix}0.1\\0.2\\0.3\end{bmatrix}}.
$$
None of the two sequences in \eqref{eq:example} converge for most starting points $x_0$.
Figure \ref{fig:no-convergence} illustrates this issue by showing the behaviour of the three coordinates of the iteration $x_k$ as in \eqref{eq:example}, for a random sampled starting point $x_0$, sampled from a uniform distribution in $[0,1]^3$.  
We observed the same behaviour for the sequence $z_k$ and for any starting point sampled this way.

\section{Proof of Theorem \ref{thm:main}.}

Consider the iteration in \eqref{eq:general_feature_spreading}.  
As $\sigma\in\hom_+(a)$ and $\varrho\in\hom_+(b)$ we have 
\begin{align*}
\Phi(\lambda F) &= D^{-1/2}KW\sigma(K^\top \varrho(\lambda D^{-1/2}F)) = D^{-1/2}KW\sigma(\lambda^b K^\top \varrho(D^{-1/2}F)) \\
&= \lambda^{ab} D^{-1/2}KW\sigma(K^\top \varrho(D^{-1/2}F)) = \lambda \Phi(F)
\end{align*}
i.e.\ $\Phi$ is one-homogeneous. Moreover, as $K$ and $D$ are nonnegative matrices, $\Phi\in\hom_+(1)$. Similarly, as every node appears in at least one hyperedge, we see that under the assumptions on $\sigma$ and $\varrho$ it holds $\varphi(F)>0$ for all $F>0$. Thus, a similar computation as the one above shows that $\varphi(F)\in \hom_+(1)$.  As a consequence, for any $F$ with $\varphi(F)=1$, every component of   $\Phi(F)$ is bounded and positive, i.e.\ there exist constants $M_{i,j}>0$ such that 
$$
\max_{F: \varphi(F)=1}\Phi(F)_{ij} = \max_F \frac{\Phi(F)_{ij}}{\varphi(F)}\leq M_{ij}\, .
$$
Hence, defining $M= \max_{ij} M_{ij} >0 $ we have that $\Phi(F)\leq M$ entrywise, for all $F$ such that $\varphi(F)=1$. As a consequence, since $U$ is entrywise positive, there exists a constant $\tilde M>0$ such that $\Phi(F)\leq \tilde M U$ entrywise,  for all $F$ such that $\varphi(F)=1$. Using Theorem 3.1 in \cite{tudisco2021nonlinear} we deduce that $F^{(k)}\to F_\star$ as $k\to \infty$ with $F_\star$ unique fixed point $F_\star= \alpha \Phi(F_\star) + (1-\alpha)U$ such that $\varphi(F_\star)=1$ and $F_\star>0$. 

Now we show that $F_\star$  is also the only point where the gradient of 
\[
\tilde \ell(F)  := \Big\|F-\frac U {\varphi(U)}\Big\|^2 + \lambda\Omega_{\mu_{\sigma,\varrho}}(F)
\]
vanishes. To this end, denoted by $S(F)$ is the $m\times (c+d)$ matrix of the hyperedge embedding $S(F) = \sigma(K^\top \varrho(F))$ and let $B(F) =KWS(F)= KW\sigma(K^\top\varrho(F))$. Notice that, with this notation, as observed in \eqref{eq:mu_e}, we can  write  
\[
\mu_{\sigma,\varrho}(\{f_i : i\in e\}) = S(F)_{e,:}\, .
\]
As a consequence we get
\[
\Omega_{\mu_{\sigma,\varrho}}(F) = \sum_{i\in V}\sum_{e:i\in e}w(e)\Big\|(D^{-1/2}F)_{i,:}- \frac 1 2 S(D^{-1/2}F)_{e,:}\Big\|^2,
\]
where, as it will be more convenient in the computation below, we are multiplying the  $\mu_e$ term in the loss by $1/2$. Of course we can always do this by rescaling one of the two mappings $\sigma$ or $\varrho$ by a factor two, without losing any of their relevant properties nor generality in the proof. Thus,  we have
\begin{align*}
\Omega_{\mu_{\sigma,\varrho}}(D^{1/2}F) &= \sum_i \sum_{e:i\in e}w(e)\sum_j(F_{ij}-\frac 1 2 S(F)_{ej})^2  \\
&=\sum_i \sum_{e:i\in e}w(e)\sum_j (F_{ij}^2-F_{ij}S(F)_{ej})  + \frac 1 4\sum_i\sum_{e:i\in e}\sum_j w(e) S(F)_{ej}^2\\
&= \sum_i \sum_j F_{ij}^2 \delta_i-  F_{ij}B(F)_{ij} + \varphi(D^{1/2}F)^2\\
&= \langle F,DF- B(F)\rangle + \varphi(D^{1/2}F)^2\, ,
\end{align*}
where $\langle\cdot,\cdot\rangle$ denotes the matrix Frobenius scalar product. 
Therefore, it holds 
\begin{align*}
    \Omega_{\mu_{\sigma,\varrho}}(F)-\varphi(F)^2 &= \langle F,F-D^{-1/2}B(D^{-1/2}F)\rangle = \langle F,F-\Phi(F)\rangle\, .
\end{align*}
As $\Phi$ is 1-homogeneous and differentiable, by the Euler theorem for homogeneous functions we have that 
\begin{align*}
    &\frac d {dF}\, \{\Omega_{\mu_{\sigma,\varrho}}(F)-\varphi(F)^2\} =\frac d {dF} \langle F,F- \Phi(F)\rangle = 2(F-\Phi(F))\, .
\end{align*}
Thus, 
\begin{align*}
    &\frac d {dF}\, \{\tilde \ell(F)-\lambda \varphi(F)^2\}   = 2 (F-U/\varphi(U) + \lambda(F-\Phi(F))  = 2( (1+\lambda)F - \lambda \Phi(F) - U/\varphi(U) )
\end{align*}
which shows that the gradient of $\tilde \ell(F)-\lambda \varphi(F)^2$ vanishes on a point $F_\star$ if and only if $F_\star$ is such that 
$$
F_\star = \frac\lambda {1+\lambda} \Phi(F_\star) + \frac {1}{1+\lambda}\frac{U}{\varphi(U)}
$$
i.e.\ $F_\star$ is a fixed point of \eqref{eq:general_feature_spreading} for $\lambda = \alpha/(1-\alpha)$ and $U = U/\varphi(U)$. 
Finally, as the two loss functions $\tilde \ell(F)$ and $\tilde \ell(F)-\lambda \varphi(F)$ have the same minimizers on the slice $\{F:\varphi(F)=1\}$, we conclude.

\section*{Acknowledgments}
This research was supported in part by ARO Award W911NF19-1-0057, ARO MURI, NSF Award DMS-1830274, and JP Morgan Chase \& Co.

\bibliography{main}
\bibliographystyle{plain}

\end{document}